\pdfoutput=1
\documentclass[11pt]{article}
\usepackage{ACL2023}
\usepackage{times}
\usepackage{latexsym}
\usepackage[T1]{fontenc}
\usepackage[utf8]{inputenc}
\usepackage{microtype}
\usepackage{arydshln}
\usepackage{amsmath}
\usepackage{inconsolata}

\usepackage{graphicx} 
\graphicspath{ {images/} }

\newcommand{\specialcell}[2][l]{%
  \begin{tabular}[#1]{@{}l@{}}#2\end{tabular}}

\title{ 
Answer Candidate Type Selection: Text-to-Text Language Model for \\Closed Book Question Answering Meets Knowledge Graphs \\

}

\author{
\textbf{Mikhail Salnikov\textsuperscript{1}}, \textbf{Maria Lysyuk}\textsuperscript{1},
\textbf{Pavel Braslavski}\textsuperscript{3},\\
\textbf{Anton Razzhigaev\textsuperscript{1,2}}, \textbf{Valentin Malykh\textsuperscript{4}}, \textbf{Alexander Panchenko\textsuperscript{1,2}} \\
\textsuperscript{1}Skolkovo Institute of Science and Technology, \textsuperscript{2}Artificial Intelligence Research Institute,\\ \textsuperscript{3}Ural Federal University, \textsuperscript{4}ISP RAS Research Center for Trusted Artificial Intelligence
\\
\url{{m.salnikov,a.panchenko}@skol.tech} 
}

\begin{document}
\maketitle
\begin{abstract}

Pre-trained Text-to-Text Language Models (LMs), such as T5 or BART 
yield promising results in the Knowledge Graph Question Answering (KGQA) task. 
However, the capacity of the models is limited and the quality decreases for questions with less popular entities.  
In this paper, we present a novel approach which works on top of the pre-trained Text-to-Text QA system to address this issue. Our simple yet effective method performs filtering and re-ranking of generated candidates based on their types derived from Wikidata \texttt{instance\_of} property. 
%
This study demonstrates the efficacy of our proposed methodology 
across three distinct one-hop KGQA 
datasets. Additionally, our approach yields results comparable to other existing specialized KGQA methods. In essence, this research endeavors to investigate the integration of closed-book Text-to-Text QA models and KGQA.

%

%
\end{abstract}

\section{Introduction}


Information stored in Knowledge Graphs (KG), such as Wikidata~\cite{wikidata}, for general domain  or some specific knowledge graphs, e.g. for the medical domain~\cite{huang2021knowledge}, can be used to answer questions in natural language.  Knowledge Graph Question Answering (KGQA) methods provide not a simple string as an answer, but instead an entity a KG.  

Pre-trained Text-to-Text LMs, such as T5~\cite{2020t5} or BART~\cite{lewis2019bart}, showed promising results on Question Answering (QA). Besides, recent studies have demonstrated the potential of Text-to-Text models to address Knowledge Graph Question Answering problems~\cite{roberts2020much,sen-etal-2022-mintaka}. 

While fine-tuning a Text-to-Text LM can significantly improve its performance, there are cases where  questions cannot be answered without  access to a knowledge graph, especially in case of less popular entities~\cite{mallen2022not}: not all required knowledge can be ``packed'' into parameters of a neural model. However, even in such cases, Text-to-Text models can generate plausible answers that often belong to the \textit{same type} as the correct answer. For example, Text-to-Text answers to the question ``What is the place of birth of Philipp Apian?'' are not correct (e.g., T5 model produced ``Neuilly-sur-Seine'' or ``Freiburg im Breisgau'' as answers), but these wrong candidates are of the correct type. Namely, the correct type ``city'' can be derived from the list of generated answers and used to perform a local KG search around the question entity ``Philipp Apian'' to derive the correct answer ``Ingolstadt''. Motivated by these observations, this study presents a method for answer type prediction utilizing the output of pre-trained Text-to-Text language models. 

The contributions of our study are as follows: (1) A simple yet effective approach for improving generative KGQA using candidate answer type selection method based on \texttt{instance\_of} properties aggregated from diversified beamsearch. (2) An open implementation of the method that is easily applicable to pre-trained generative models.\footnote{\url{https://github.com/s-nlp/act}}

\section{Related Work}

Traditional KGQA methods can be classified into two categories: retrieval-based and semantic parsing. Retrieval-based methods involve vectorizing the textual question and projecting it into a graph-based vector space containing candidate entities \cite{Huang2019KnowledgeGE, m3m}. Semantic parsing approaches generate formal question representations (e.g., SPARQL queries) to query a KG for the answer. 
Retrieval-based approaches rely on computationally expensive similarity searches using vector indices of millions of candidate entities. Semantic parsing requires maintaining a graph database capable of processing SPARQL queries.

Recently, to address these shortcomings of existing methods, a third wave of approaches emerged based on pre-trained Text-to-Text LMs such as T5 \cite{2020t5} or BART \cite{lewis2019bart}. Given a question, these models generate a label of the answer that can be directly linked to the entity in a KG. These models are more computationally convenient and they are described below.

The \emph{Text-To-Text Transfer Transformer (T5)} \cite{2020t5} is effective for question answering, as demonstrated by \citet{roberts2020much}, or as part of a retrieval pipeline \cite{izacard2020leveraging}. Furthermore, it has been shown that training T5 with Salient Span Masking (SSM) improves the model's performance on QA task.
T5-ssm involves tuning T5 as a language model, masking \emph{entities} instead of random tokens. 
T5-ssm-nq is a variant of the T5-ssm that is additionally fine-tuned on the NaturalQuestions~(NQ) \cite{kwiatkowski2019natural} dataset. \emph{BART}, a Text-to-Text model trained as a denoising autoencoder~\cite{lewis2019bart}, can also be applied to KGQA task~\cite{KQAPro}.


\section{Answer Candidate Type Selection}

\begin{figure}[h]
    \centering
    \includegraphics[width=0.45\textwidth]{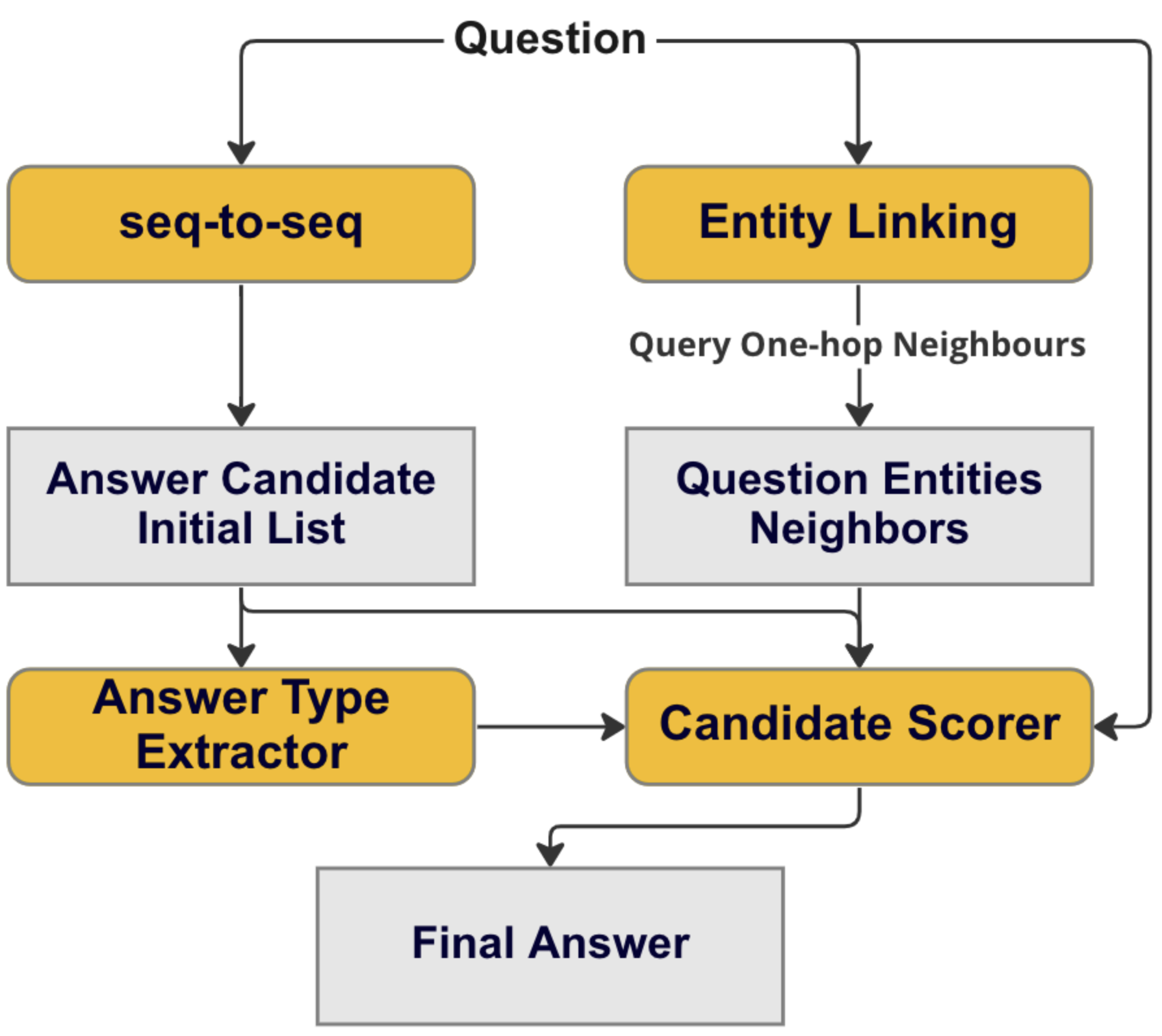}
    \caption{Answer Candidate Type (ACT) Selection. }
    \label{fig:pipeline}
\end{figure}

\begin{figure*}[h]
    \centering
    \includegraphics[width=0.99\textwidth]{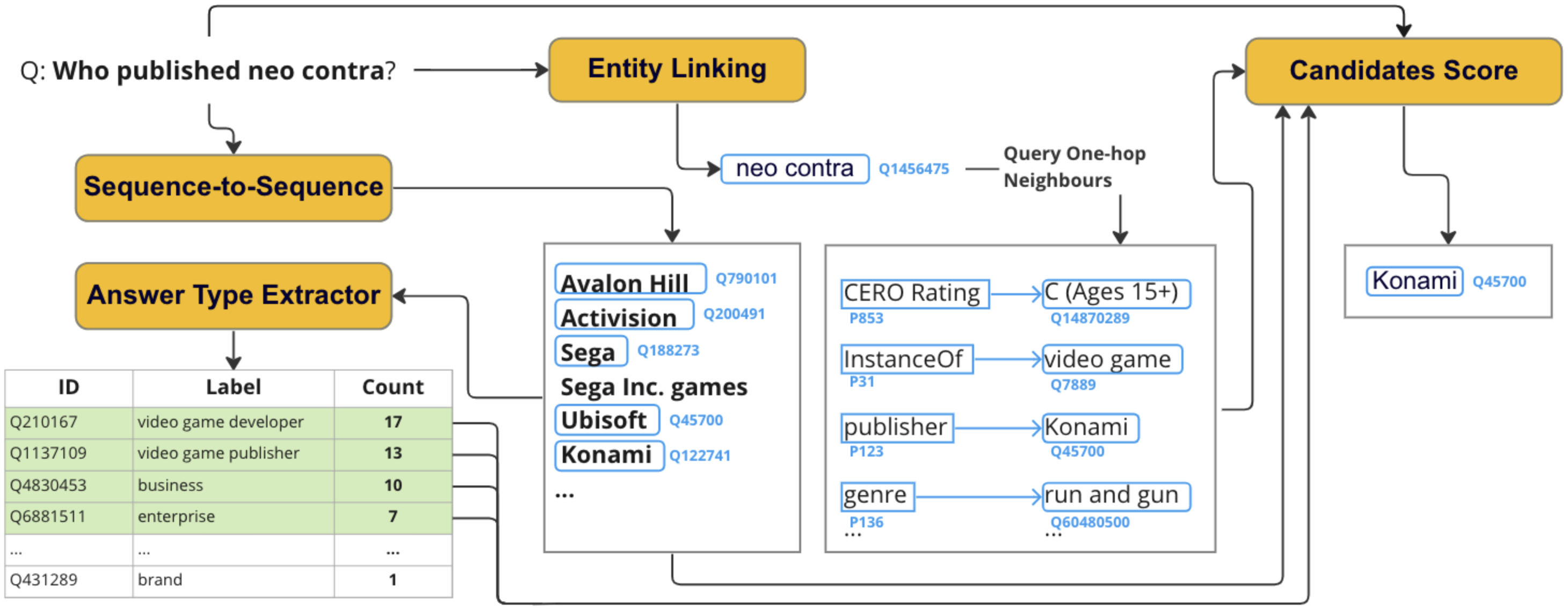}
    \caption{An example of the proposed Answer Candidate Type (ACT) Selection result. }
    \label{fig:pipeline_example}
\end{figure*}

This section presents our proposed approach, Answer Candidate Type (ACT) Selection.
We propose a universal approach to selecting the correct answer in the KGQA task by using any pre-trained sequence-to-sequence (seq2seq) model (in our cases a Text-to-Text Language Model) 
to generate answer candidates and to infer the type of expected answer. The answer candidate type selection pipeline shown in Figures~\ref{fig:pipeline} and~\ref{fig:pipeline_example} consists of four parts: the Text-to-Text model for candidate generation, Answer Type Extractor, Entity Linker, and the Candidate Scorer.

\subsection{Initial Answer Candidate List Generation}

To increase the diversity of the generated results, we use Diverse Beam Search~\cite{vijayakumar2016diverse} to generate an initial list of answer candidates $C$. It often leads to a better exploration of the search space by ensuring that alternative answers are considered. We define the types of entities using the Wikidata property \texttt{instance\_of}~(P31). Note that an entity can be of multiple types. Finally, the initial list of answer candidates is used in the Answer Candidate Typing and the Candidate Scorer with the mined candidates. 


\subsection{Answer Candidate Typing} \label{act}

We rank all types by their frequency in the initial list of answer candidates. 
After that, we merge the top-$K$ most frequent types and similar types to the final list $T$.
Types similarity is calculated as a cosine similarity between Sentence-BERT~\cite{reimers-2019-sentence-bert} embeddings of respective labels. The final types are defined as the ones where similarity is greater than a threshold.

A similar aggregation method using hypernyms (also known as ``is-a''  or ``instance-of'' relations) was used in the past to label clusters of words senses in distributional models~\cite{biemann2013text,panchenko-etal-2017-unsupervised-mean}: distributionally similar words share common hypernym and ``top'' common hypernyms are surprisingly good labels for sense clusters. The analogy in our method is that Text-to-Text models appear to produce a list of distributionally similar candidates.  

\subsection{Entity Linking}

To enrich the list of candidates, we add all one-hop neighbours of the entities found in the question. For that we use the fine-tuned spaCy Named Entity Recognition~(NER)\footnote{\url{https://spacy.io}. More details about  fine-tuning of the NER can be found in Appendix~\ref{sec:appendix:ner}.} and the mGENRE~\cite{de2022multilingual} entity linking model. 

\subsection{Candidates Scorer}
Finally, we calculate four scores for each 
answer candidate and rank them based on the weighted sum of the scores.
The scores are as follows:

\textbf{(1)~Type score} represents the size of the intersection between the set of types extracted from the answer candidates and the selected answer types. It is weighted by the number of selected answer types: $$S_\textrm{type}~=~\frac{|\textrm{Candidates' Types} \cap T|}{|T|}.$$

\textbf{(2)~Forward one-hop neighbors score} $S_\textrm{neighbour}$ is  assigned 1 if the candidate is among the neighbors of the question entities, and 0 otherwise.

\textbf{(3)~Text-to-Text answer candidate score} is determined by the rank of the candidate in the initial list $C$ generated by the Text-to-Text model divided by the size of the list: $$S_\textrm{t2t}~=~\frac{C.\textsc{}{index}(\textrm{Candidate})}{|C|}.$$

\textbf{(4)~Question-Property Similarity score} $S_\textrm{property}$ measures the cosine similarity between the embeddings of the relevant property and the entire question. We employ Sentence-BERT~\cite{reimers-2019-sentence-bert} to encode the question, following a similar approach used for the Answer Candidate Type module.




The four scores are calculated for each entity and then are combined to generate a final score that determines the entity's ranking. The answer with the highest weighted sum of scores in the candidate list is selected as the final answer:
$$S_\textrm{final} = S_\textrm{type} + S_\textrm{neighbour} + S_\textrm{t2t} + S_\textrm{property}.$$



\section{Experiments}


We fine-tuned the Text-to-Text and spaCy NER models by using the entire training part of the respective datasets and fitting the model for eight epochs. The initial answer candidate lists were generated using Diverse Beam Search with 200 beams and a diversity penalty of 0.1. The Answer Candidate Typing module utilized the top-3 types and a similarity threshold of 0.6. 

\subsection{Data}
We evaluate the ACT Selection on three Wikidata datasets containing one-hop questions.
\emph{SimpleQuestions-Wikidata (SQWD)}~\cite{SQ_WD} is a mapping of SimpleQuestions~\cite{simplequestions} to Wikidata containing 21,957 questions. 
\emph{RuBQ}~\cite{korablinov2020rubq,rybin2021rubq} is a KGQA dataset that contains 2,910 Russian questions of different types along with their English translations. \emph{Mintaka}~\cite{sen-etal-2022-mintaka} is a multilingual KGQA dataset composed of 20,000 questions of different types. For our experiments we took only \emph{generic} questions, whose entities are one hop away from the answers' entities in Wikidata, which resulted in 1,757
English questions.

\subsection{Evaluation} \label{evaluation}

We hypothesize that even if a closed-book QA text-to-text model returns an incorrect answer, the odds are that it is of the correct type. 

The present study involves the extraction of answer types from Text-to-Text generated answers, followed by a comparison with the ground-truth answer types in the SQWD dataset. Our experimental findings demonstrate that the fine-tuned T5-Large-SSM model equipped with the ACT~Selection can accurately predict the correct answer type in \textbf{94\%} of the cases, while only \textbf{61\%} of the candidate answers share the same type as the correct answer. These results have provided an impetus to leverage this information to facilitate question-answering.

\begin{figure}[h]
    \centering
    \includegraphics[width=0.5\textwidth,keepaspectratio]{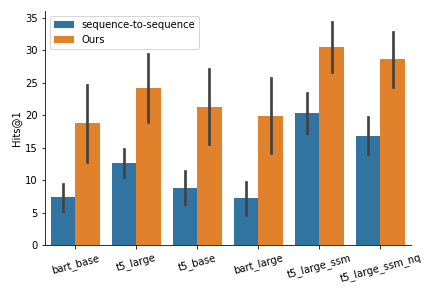}
    \caption{Average Hit@1 scores for the tuned models on SQWD, RuBQ, and Mintaka datasets from Table~\ref{tab:full_results}. 
    }
    \label{fig:mean_tuned_all_datasets}
\end{figure}

\begin{table}
\centering
  \scalebox{0.87}{
      \begin{tabular}{lcc}
        \hline
        Model & SQWD & RuBQ en \\
        \hline
        QAnswer & 33.31 & 32.30 \\
        KEQA TransE PTBG & \textbf{48.89} & 33.80 \\
        ChatGPT & 15.32 & 36.53 \\ \hline
        T5-Large-ssm (fine-tuned) & 23.66 & 21.44 \\ 
        \text{Ours}: T5-Large-ssm (fine-tuned) & 47.42 & 26.02 \\ \hline 
        T5-11b-ssm-nq (zero-shot) & 10.94 & 33.38 \\
        \text{Ours}: T5-11b-ssm-nq (zero-shot) & 38.51 & \textbf{38.31} \\
        \hline
      \end{tabular}
  }
  \caption{Comparsion of the ACT Selection with KGQA baselines in terms of Hit@1 for SimpleQuestion-Wikidata (SQWD) with T5-Large-ssm fine-tuned on its training part and T5-11b-ssm-nq in zero-shot mode.}
  \label{tab:comparsion_hits1_sqwd}
\end{table}

\begin{table*}[ht]
\centering
\resizebox{0.99\textwidth}{!}{
\begin{tabular}{lrrrrrrrrr}
\hline
             & \multicolumn{3}{c}{\textit{SimpleQuestions-Wikidata}}                                                                                                                                                           & \multicolumn{3}{c}{\textit{RuBQ (English)}}                                                                                                                                                                    & \multicolumn{3}{c}{\textit{Mintaka (one-hop, English)}}                                                                                                                                                         \\ 
                       Tuned on $\rightarrow$ & \textbf{\begin{tabular}[c]{@{}l@{}}Zero-shot\end{tabular}} & \textbf{\begin{tabular}[c]{@{}l@{}}SQWD\end{tabular}} & \textbf{\begin{tabular}[c]{@{}l@{}}Mintaka\end{tabular}} & \textbf{\begin{tabular}[c]{@{}l@{}}Zero-shot\end{tabular}} & \textbf{\begin{tabular}[c]{@{}l@{}}SQWD\end{tabular}} & \textbf{\begin{tabular}[c]{@{}l@{}}Mintaka\end{tabular}} & \textbf{\begin{tabular}[c]{@{}l@{}}Zero-shot\end{tabular}} & \textbf{\begin{tabular}[c]{@{}l@{}}SQWD\end{tabular}} & \textbf{\begin{tabular}[c]{@{}l@{}}Mintaka\end{tabular}} \\ \hline
BART-base              & 0                                                                      & 16.54                                                            & 7.08                                                                & 0                                                                      & 5.93                                                             & 3.72                                                                & 0                                                                      & 2.06                                                             & 9.12                                                                \\ 
Ours       & 30.38                                                                  & \textbf{42.60}                                                   & 30.70                                                               & 9.50                                                                   & 11.65                                                            & \textbf{11.72}                                                      & 4.70                                                                   & 5.88                                                             & \textbf{10.29}                                                      \\ \hdashline
BART-large             & 0                                                                      & 16.97                                                            & 3.02                                                                & 0                                                                      & 4.07                                                             & 4.86                                                                & 0                                                                      & 1.76                                                             & 12.65                                                               \\ 
Ours      & 30.42                                                                  & \textbf{42.64}                                                   & 31.39                                                               & 9.50                                                                   & 12.15                                                            & \textbf{12.79}                                                      & 4.41                                                                   & 5.29                                                             & \textbf{15.29}                                                      \\    \hdashline
T5-base                & 0                                                                      & 21.26                                                            & 6.19                                                                & 0                                                                      & 6.22                                                             & 6.93                                                                & 0                                                                      & 4.41                                                             & 8.24                                                                \\
Ours        & 30.47                                                                  & \textbf{43.13}                                                   & 34.60                                                               & 9.44                                                                   & 14.44                                                            & \textbf{16.58}                                                      & 4.71                                                                   & 8.53                                                             & \textbf{10.59}                                                      \\    \hdashline
T5-large               & 0                                                                      & 22.36                                                            & 9.43                                                                & 0                                                                      & 11.15                                                            & 12.15                                                               & 0                                                                      & 7.06                                                             & 14.41                                                               \\ 
Ours       & 29.88                                                                  & \textbf{43.05}                                                   & 36.89                                                               & 9.44                                                                   & 18.94                                                            & \textbf{20.51}                                                      & 4.71                                                                   & 10.00                                                            & \textbf{15.88}                                                      \\    \hdashline
T5-large-ssm           & 0.57                                                                   & 23.66                                                            & 5.92                                                                & 0.42                                                                   & 21.44                                                            & 23.87                                                               & 0.50                                                                   & 19.71                                                            & 27.65                                                               \\ 
Ours    & 23.39                                                                  & \textbf{47.42}                                                   & 36.54                                                               & 9.72                                                                   & 26.02                                                            & \textbf{27.88}                                                      & 6.76                                                                   & 18.53                                                            & \textbf{28.24}                                                      \\    \hdashline
T5-large-ssm-nq        & 5.12                                                                   & 22.52                                                            & 4.34                                                                & 18.87                                                                  & 17.80                                                            & 19.23                                                               & 17.65                                                                  & 14.12                                                            & 23.24                                                               \\ 
Ours & 35.09                                                                  & \textbf{43.88}                                                   & 36.39                                                               & \textbf{27.52}                                                         & 25.38                                                            & 26.38                                                               & 22.94                                                                  & 14.12                                                            & \textbf{25.59}                                                      \\    \hdashline
T5-11b-ssm             & 1.81                                                                   & ---                                                              & ---                                                                 & 14.09                                                                  & ---                                                              & ---                                                                 & 20.88                                                                  & ---                                                              & ---                                                                 \\ 
Ours      & \textbf{25.84}                                                         & ---                                                              & ---                                                                 & \textbf{20.94}                                                         & ---                                                              & ---                                                                 & \textbf{24.71}                                                         & ---                                                              & ---                                                                 \\    \hdashline
T5-11b-ssm-nq          & 10.94                                                                  & ---                                                              & ---                                                                 & 33.38                                                                  & ---                                                              & ---                                                                 & 41.76                                                                  & ---                                                              & ---                                                                 \\ 
Ours  & \textbf{38.51}                                                         & ---                                                              & ---                                                                 & \textbf{38.31}                                                         & ---                                                              & ---                                                                 & \textbf{45.00}                                                         & ---                                                              & ---                                                                 \\ \hline
\end{tabular}
}
\caption{Evaluation results on three one-hop KGQA datasets (Hit@1 scores): comparing Text-To-Text Language Model with and without our proposed ACT Selection approach in zero-shot (without tuning for QA) or tuned on SQWD or Mintaka.}
\label{tab:full_results}
\end{table*}


We evaluate the performance of two commonly used architecture types, T5 and BART. The proposed approach consistently improves the results of the Text-to-Text models on various datasets, as illustrated in Figure~\ref{fig:mean_tuned_all_datasets}. We compare the mean Hit@1 scores of the tuned Text-to-Text models with the aforementioned datasets. Text-to-Text models were fine-tuned on the train splits of SQWD and the full train split of Mintaka datasets, and subsequently evaluated on the test splits of SQWD, RuBQ, and Mintaka using both tuned versions of the models. 

As demonstrated in Table~\ref{tab:full_results}, the proposed approach consistently enhances the quality of KGQA tasks across various Text-to-Text models. Furthermore, we conducted experiments to verify that the proposed method can be employed with the Text-to-Text models in a zero-shot learning manner, without any fine-tuning. The benefits of the approach, in terms of quality improvement, are more noticeable when applied to smaller models. For example, the T5-large model, with its 737 million parameters, when paired with ACT Selection, delivers comparable performance to the T5-11b model, which has 11 billion parameters.

In line with expectations, larger models generally yield superior results. Notably, T5 models using the suggested method outperformed BART models. Moreover, across all tested T5 and BART models, implementing the ACT Selection markedly enhanced the performance of the foundational Text-to-Text model.

Table~\ref{tab:comparsion_hits1_sqwd} showcases performance comparison between our suggested method and prominent KGQA systems, namely QAnswer~\cite{diefenbach2020towards}, KEQA~\cite{Huang2019KnowledgeGE}, and chatGPT.\footnote{\url{https://openai.com/blog/chatgpt}} QAnswer is a multilingual rule-based system that tranforms the question into a SPARQL query. KEQA utilizes TransE embeddings of 200 dimensions, trained on Wikidata using the Pytorch-BigGraph (PTBG) framework~\cite{pbg}. ChatGPT is a conversational model that was launched in late 2022 and has received worldwide acclaim. Further details about evaluating ChatGPT and other generative models through entity-linked predictions can be found in appendix~\ref{sec:appendix:eval_gen_model}. The tabulated data reveals that our approach delivers outcomes commensurate with those of state-of-the-art (SOTA) systems.

\subsection{Ablation Study}


\begin{table*}[ht]
  \footnotesize
  \centering
  \begin{tabular}{lrrrrr}
    \hline
    & Type score & \specialcell{Forward one-hop\\neighbours score} & \specialcell{Text-to-Text LM\\candidates score} & \specialcell{Question-Property\\Similarity score} & All scores \\
    \hline
    \specialcell{Only initial candidates\\generated by Text-to-Text} & 2.51 & 31.73 & 27.04 & 31.82 & 35.89 \\
    \hline
    \specialcell{Only question\\neighbours candidates} & 5.07 & 4.84 & 4.52 & 29.86 & 30.06 \\
    \hline
    \specialcell{Full answer\\candidates set} & 2.81 & 5.46 & 27.04 & 30.75 & \textbf{47.42} \\
    \hline
  \end{tabular}
  \caption{Ablation study of ACT Selection. Reporting Hit@1 at SQWD for T5-large-ssm fine-tuned on SQWD.}
  \label{tab:ablation_study}
\end{table*}



We conducted an ablation study (cf. Table~\ref{tab:ablation_study}) to investigate the effects of the proposed scores on the candidate set collection process. Our main goal was to confirm that incorporating type information enhances candidate selection. We observed that methods relying solely on scores (such as Question-Property Similarity score) were not as effective as the ACT~Selection approach.

Furthermore, we examined the necessity of initial candidates generated by the Text-to-Text model and whether restricting to question entity neighbors was sufficient. This investigation aimed to determine the added value of initial candidates in the selection process.



\subsection{Error Analysis} 

We showed above that the ACT~Selection approach fixed errors produced by the Text-to-Text LMs. We evaluate this approach using a subset of questions and predictions from the T5-Large-SSM model for the SQWD dataset. Our focus is on questions where the model's top-1 prediction was incorrect, but the ACT~Selection approach extracted the correct answer.

The Text-to-Text model generated the correct answer in only 58.4\% of questions in the chosen subset. However, our Entity Linking module was able to correctly extract 99.11\% of question entities for this subset. The extraction of additional candidates from the question entity neighbors played a critical role in finding the correct answer.



\section{Conclusion} 

We introduced a method for question answering over knowledge graph based on post-processing of beam-search outputs of a Text-to-Text model. Namely, a simple aggregation of KG ``instance-of'' relations is used to derive a likely type of the answer. This simple technique consistently improves performance of various Text-to-Text LMs favorably comparing to both specialized KGQA methods and ChatGPT with a carefully selected prompt and entity linked output on three distinct English one-hop KGQA datasets.

Our method may be also used to directly perform answer typing. In principle, it can be straightforwardly adapted to multilingual setup, but also multi-hop questions.
We find it promising to use the method with larger pre-trained models to further boost performance as our current experiments show that the QA quality grows with the model size. 



\section{Limitations}
The main limitation of the current study is that the approach was only tested for one-hop questions. In principle, one can, however, sample candidates from graph from arbitrary subgraphs, e.g. second-order ego-networks of entity found in question. At the same time, improvements shown in this paper may not nessesarily generalize to such setting and need to be tested.

Another limitation is using diverse beam search, which is a computationally more  expensive process as it requires larger beam sizes, usually.

Finally, requesting KG data can be a bottleneck if one is using a public SPARQL endpoint with query limits. This limitation can be alleviated by using an in-house private copy of a KG.

\section{Ethical Considerations}
Large pre-trained Text-to-Text models such as those used in our work are trained on datasets which may contain biased opinions. Therefore, QA/KGQA systems built on top of such models may transitively reflect such biases potentially generating stereotyped answers to the questions. As a consequence, it is recommended in production, not research settings, to use a special version of debiased pre-trained neural models and/or other technologies for the alleviation of the undesired biases of LLMs.

\bibliography{main, anthology}
\bibliographystyle{acl_natbib}

\appendix
\section{Named Entity Recognition}
\label{sec:appendix:ner}
According to the recent review of SOTA NER \cite{vajjala2022we}, top-3 approaches were chosen: spaCy\footnote{\url{https://spacy.io}}, Stanza\footnote{\url{https://stanfordnlp.github.io/stanza/}} and SparkNLP\footnote{\url{https://nlp.johnsnowlabs.com}}. Pre-rained NERs showed very poor quality ranging from 64\% to 88\% of missing cases for the SQWD data set. Among them, spaCy was the best; therefore, the standard spaCy configuration\footnote{\url{https://spacy.io/usage/training/}} was chosen for further fine-tuning. This pipeline requires two main pre-processing steps. First, the span of the entity should be fed into the algorithm. This span is predefined for Mintaka. However, for SQWD and RuBQ only Wikidata IDs of the entities are presented. Therefore, it was necessary first to define labels of the entities and all corresponding redirects. Next, these labels should have been found in the initial sentence for the span detection. Since for some of the entities there was no direct match in the sentence, the fuzzy search\footnote{\url{https://pypi.org/project/fuzzywuzzy/}} was started. Second, spaCy requires the tag of the entity label (e.g., PERSON for Elon Musk , ORG for Tesla - the so-called BIO type tagging) for training, but in the initial data this label is missing. PERSON tag was chosen as the one for all cases. Additional experiments with partial data tagging (defining exact tag for each entity) were not successful.

\section{Evaluation generative models on KGQA problem}
\label{sec:appendix:eval_gen_model}
To link predicted answers with entities, we utilized the full-text search engine provided by the Wikidata API\footnote{\url{https://www.wikidata.org/w/api.php}}. For answers generated by ChatGPT, we performed an additional step of removing the trailing dot at the end of the prediction (e.g., changing `Yes.' to `Yes'). For RuBQ dataset we just checked that predicted entity is one of the possible answers. 

For predicting answers in the KGQA style, we experimented with different prompts for ChatGPT. Specifically, we used the prompt `Answer as briefly as possible without additional information.' for evaluating the SQWD dataset and `Answer as briefly as possible. The answer should be `Yes', `No' or a number if I am asking for a quantity of something, if possible, otherwise just a few words.' for the RuBQ dataset.

\section{Examples}
In this section, we include figures that illustrate examples of the working pipeline. Figure~\ref{fig:pipeline_example} presents the pipeline for the question "Who published neo contra?" The Text-to-Text model generates a set of answer candidates, such as "Avalon Hill," "Activision," and "Sega." These candidates are used to extract the type information, such as "video game developer." This type information is then employed in the Candidate Score module to rerank the final set of candidates, ultimately identifying the correct answer as "Konami."

\begin{figure*}[h]
    \centering
    \includegraphics[width=0.99\textwidth,keepaspectratio]{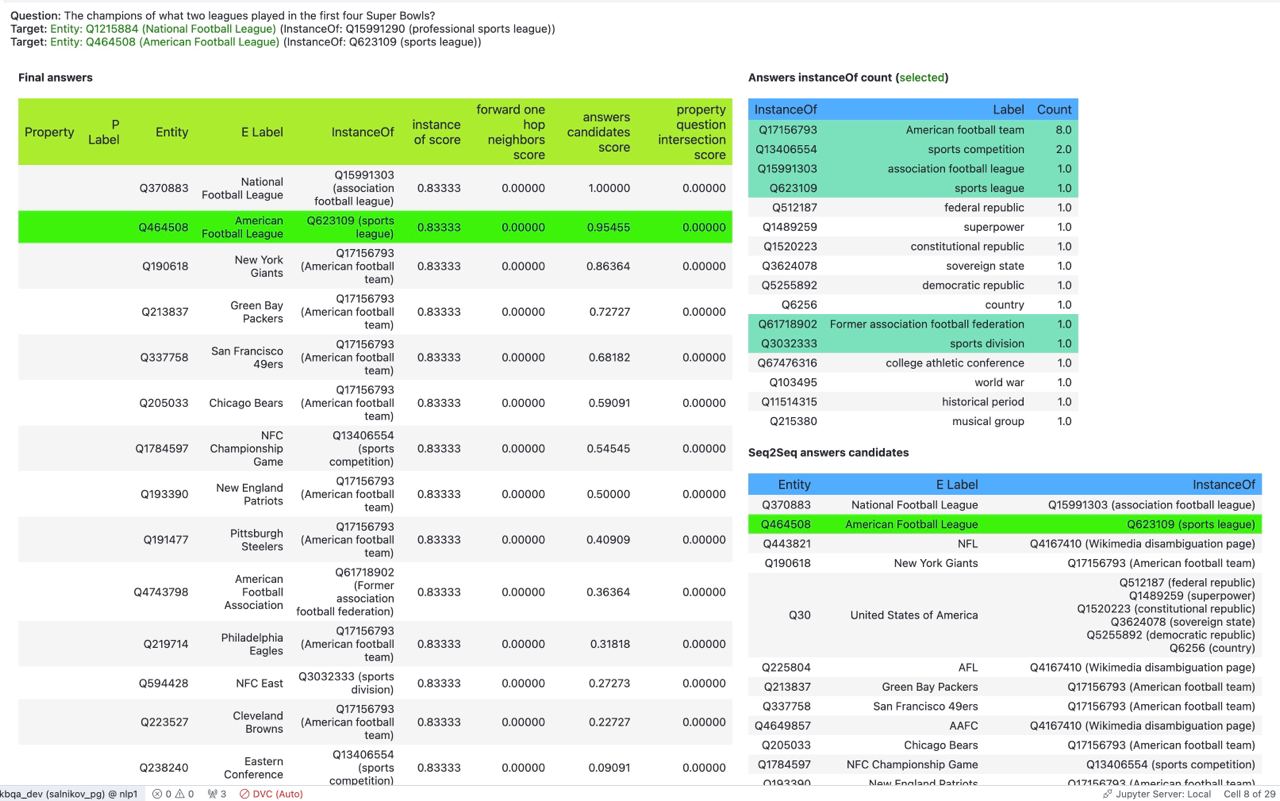}
    \caption{Example question: The champions of what two leagues played in the first four Super Bowls?}
    \label{fig:appendix_example_1}
\end{figure*}

\begin{figure*}[h]
    \centering
    \includegraphics[width=0.99\textwidth,keepaspectratio]{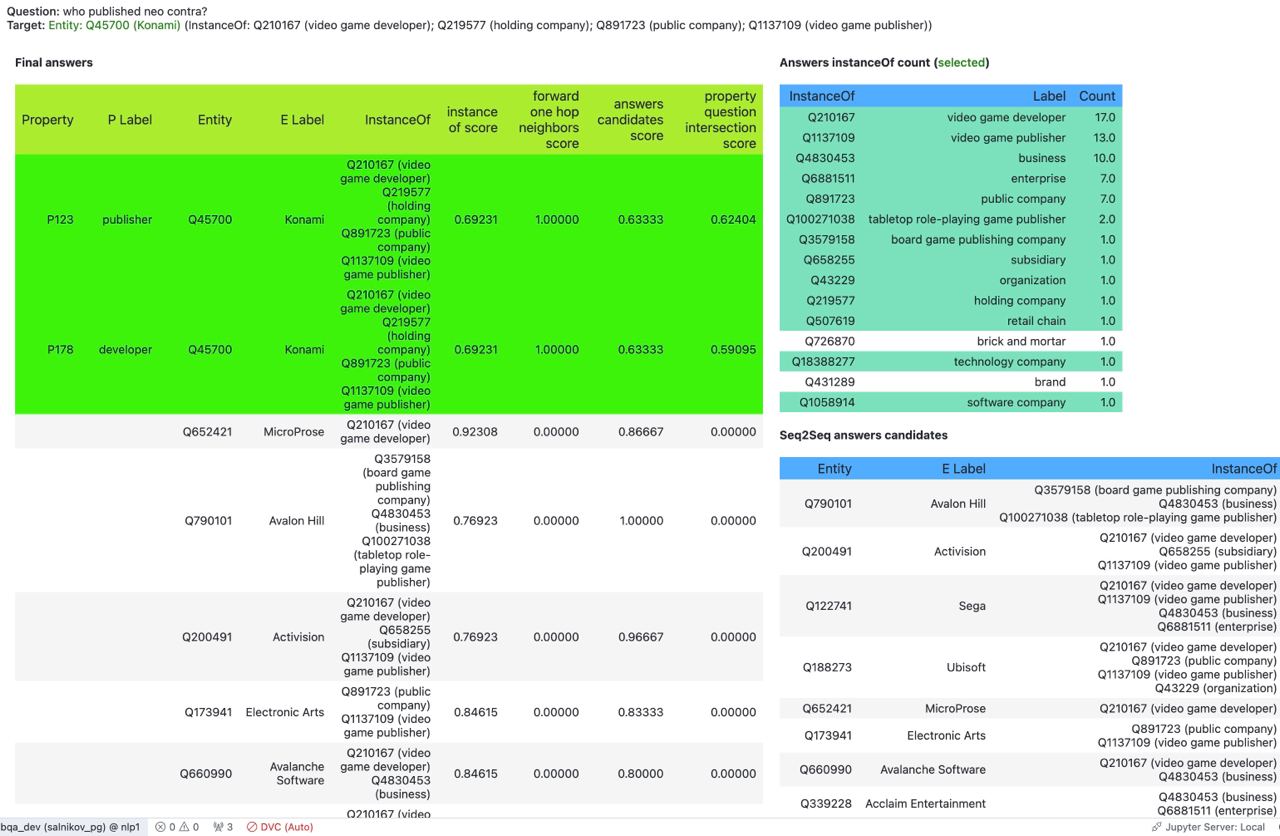}
    \caption{Example question: Who published neo contra?}
    \label{fig:appendix_example_2}
\end{figure*}

\begin{figure*}[h]
    \centering
    \includegraphics[width=0.99\textwidth,keepaspectratio]{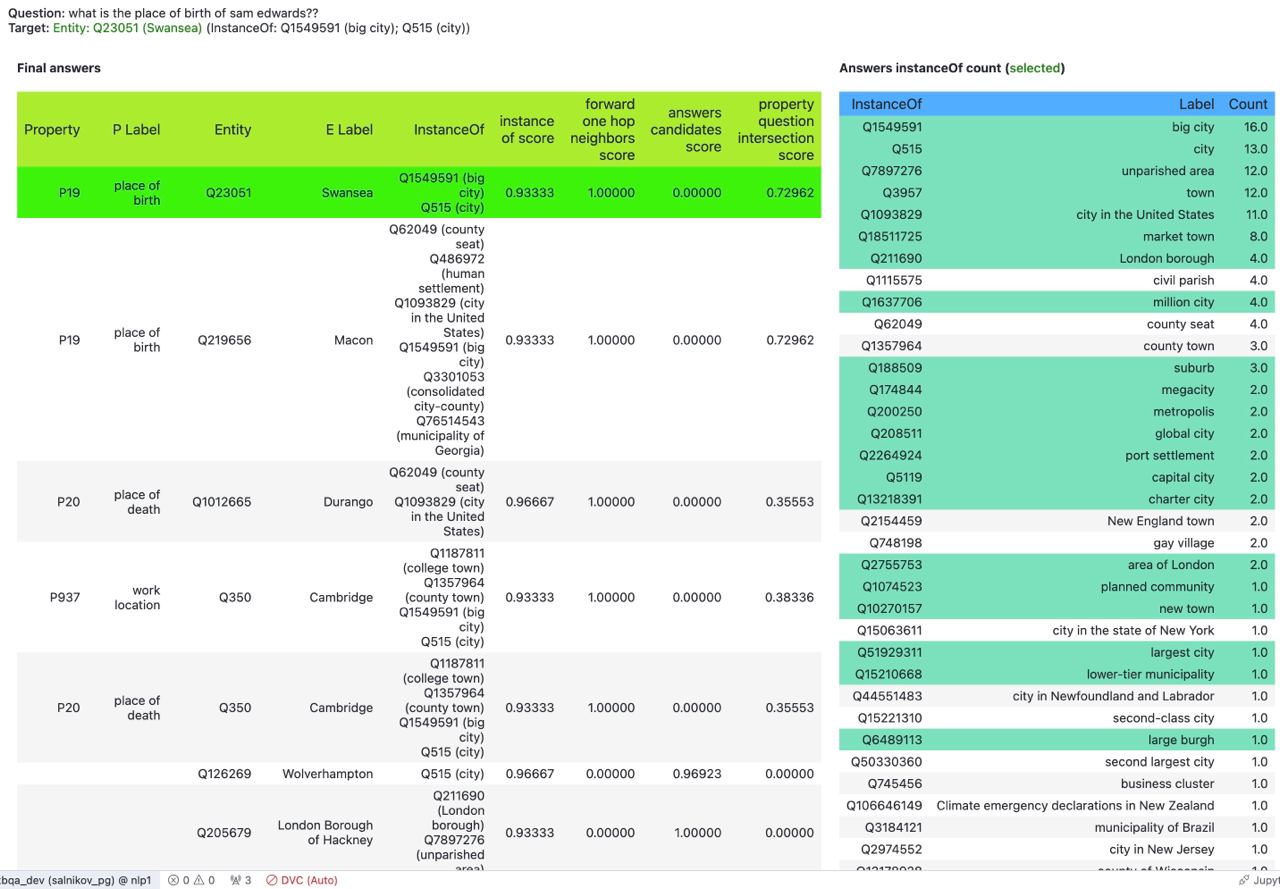}
    \caption{Example question: What is the place of birth of Sam Edwards?}
    \label{fig:appendix_example_3}
\end{figure*}

Additionally, in Figures~\ref{fig:appendix_example_1}, \ref{fig:appendix_example_2}, and  \ref{fig:appendix_example_3}, we provide additional examples that demonstrate the extraction of types and the calculation of scores within the pipeline.

\end{document}